%% file: paper.tex
\pdfoutput=1

\documentclass[11pt]{article}

\usepackage[]{acl}

\usepackage{times}
\usepackage{latexsym}
\usepackage[T1]{fontenc}
\usepackage[utf8]{inputenc}
\usepackage{microtype}
\usepackage{inconsolata}
\usepackage{booktabs}
\usepackage{graphicx}
\usepackage{multirow}
\usepackage{xspace}
\usepackage{float}
\usepackage{placeins}

\usepackage{hyperref}
\usepackage{ragged2e}
\usepackage{tabularx}
\usepackage[most]{tcolorbox}
\usepackage{xcolor}

\usepackage{soul}

\newcommand{\Vectara}{Vectara\xspace}
\newcommand{\hallucinationleaderboard}{original hallucination leaderboard\xspace}
\newcommand{\hhem}{HHEM\xspace}
\newcommand{\faithjudge}{FaithJudge\xspace}
\newcommand{\faithbench}{FaithBench\xspace}

\usepackage[hang,flushmargin]{footmisc}
\title{Benchmarking LLM Faithfulness in RAG with Evolving Leaderboards
}

\author{
\bf Manveer Singh Tamber$^{1}$\thanks{Work done while at \Vectara.}, Forrest Sheng Bao$^{3*}$, Chenyu Xu$^{2, 3}$, Ge Luo$^{2}$, Suleman Kazi$^{2}$, \\  \bf  Minseok Bae$^{4*}$, Miaoran Li$^{3*}$,  Ofer Mendelevitch$^2$, Renyi Qu$^{2}$, Jimmy Lin$^1$
\\[1ex]
$^1$~University of Waterloo \quad $^2$~Vectara \quad  
 $^3$~Iowa State University  \quad $^4$~Stanford University 
\\[1ex]
        \textbf{Correspondence:} \texttt{\{mtamber, jimmylin\}@uwaterloo.ca}, \\
        \texttt{ofer@vectara.com}, \texttt{forrest.bao@gmail.com}\\
}

\begin{document}
\maketitle

\begin{abstract}
Retrieval-augmented generation (RAG) aims to reduce hallucinations by grounding responses in external context, yet large language models (LLMs) still frequently introduce unsupported information or contradictions even when provided with relevant context. 
This paper presents two complementary efforts at \Vectara to measure and benchmark LLM faithfulness in RAG.
First, we describe our \hallucinationleaderboard, which has tracked hallucination rates for LLMs since 2023 using our \hhem hallucination detection model.
Motivated by limitations observed in current hallucination detection methods, we introduce \faithjudge, an LLM-as-a-judge framework that leverages a pool of diverse human-annotated hallucination examples to substantially improve the automated hallucination evaluation of LLMs.
We introduce an enhanced hallucination leaderboard centered on \faithjudge that benchmarks LLMs on RAG faithfulness in summarization, question-answering, and data-to-text generation tasks.
\faithjudge enables a more reliable benchmarking of LLM hallucinations in RAG and supports the development of more trustworthy generative AI systems:\ \url{https://github.com/vectara/FaithJudge}.

\end{abstract}

\section{Introduction}

Large language models (LLMs) excel in various tasks, but frequently produce hallucinations, generating false or misleading information unsupported by provided contexts or world knowledge~\cite{hallucinationsurvey,hallucinationsurvey2, truthfulqa, aggrefact}.
While retrieval-augmented generation (RAG) approaches~\cite{pmlr-v119-guu20a, rag,shuster-etal-2021-retrieval-augmentation} seek to mitigate hallucinations by grounding responses in trusted contexts, they do not fully eliminate hallucinations, as LLMs often introduce details unsupported by retrieved contexts, misrepresent information, or generate outright contradictions~\cite{ragtruth}.

An ongoing challenge within RAG is evaluating and ensuring context-faithfulness~\cite{ragtruth, jia-etal-2023-zero, ming2024faitheval}.
In this paper, we mainly focus on evaluating faithfulness in summarization tasks, building upon extensive prior research on summary consistency evaluation.
Summarization tasks provide a practical benchmark for faithfulness, thanks to rich available hallucination datasets and established automated evaluation methods.
However, despite recent progress, both fine-tuned detection models and LLM-as-a-judge techniques~\cite{llmjudge,luo2023chatgpt,jacovi2025facts} continue to struggle with accurately identifying hallucinations in LLM outputs.

We present two complementary efforts at \Vectara for measuring and benchmarking LLM faithfulness in RAG. First, we describe our \hallucinationleaderboard, which since 2023 has tracked hallucination rates for LLMs, currently ranking over 160 LLMs on hallucinations in summarization tasks.
Second, motivated by the limitations of current hallucination detection approaches, we introduce \faithjudge, an LLM-as-a-judge framework that leverages a pool of diverse, human-annotated hallucination examples to improve automated faithfulness evaluation. 

\faithjudge leverages labelled hallucination annotations from diverse LLM generations to 
automate the evaluation of LLMs on their propensity to hallucinate when summarizing the same articles or using the same articles to respond to queries.
This approach results in notably higher agreement with human judgments compared with existing automated methods.

While our main investigation focuses on summarization tasks, we expand \faithjudge to other RAG tasks (including QA and data-to-text generation) using RAGTruth~\cite{ragtruth}, further detailed in Section~\ref{ragtruth_faithjudge}.

Unlike previous work in the automated evaluation of LLMs for faithfulness in RAG,  \faithjudge recasts LLM hallucination evaluation:\ the judge LLM studies a handful of human-annotated peer responses to the same source and then rules on a fresh candidate.
Because the judge learns directly from the examples in its prompt, it needs no extra training, adapts naturally to new domains and tasks given annotations, and reaches state-of-the-art agreement with human annotators compared to existing methods while remaining fully automated.

\section{Background}

Numerous datasets have been developed for evaluating hallucinations in summarization tasks.
Earlier datasets, such as SummaC~\cite{laban-etal-2022-summac} and AggreFact~\cite{aggrefact}, aggregated multiple resources and standardized labels and classification taxonomies.
However, these primarily focused on summaries from pre-ChatGPT models like fine-tuned T5~\cite{t5}, BART~\cite{lewis-etal-2020-bart}, and PEGASUS~\cite{pegasus}, potentially limiting their relevance to contemporary LLMs that may produce more nuanced and difficult-to-identify hallucinations.

Recent benchmarks address this limitation by incorporating summaries generated by modern LLMs.
TofuEval~\cite{tang-etal-2024-tofueval} provided hallucination labels on topic-focused dialogue summarization tasks with LLMs including GPT-3.5-Turbo, Vicuna~\cite{vicuna2023} and WizardLM~\cite{xu2023wizardlm}.
Similarly, HaluEval~\cite{halueval} included ChatGPT-generated hallucinations across summarization, question-answering (QA), and dialogue tasks, while RAGTruth~\cite{ragtruth} also annotated responses from models including GPT-3.5, GPT-4~\cite{gpt4}, Llama-2~\cite{touvron2023llama}, and Mistral~\cite{jiang2023mistral7b}.
\faithbench~\cite{bao-etal-2025-faithbench} presented human annotations of challenging hallucinations in summaries from 10 modern LLMs from 8 different model families (detailed further in Section~\ref{sec:faithbench_main}).

Due to limited large-scale, human-annotated data for training hallucination detectors, early detection methods relied heavily on natural language inference (NLI) or question-answering (QA) systems~\cite{fabbri-etal-2022-qafacteval}.
For instance, SummaC aggregated sentence-level NLI entailment scores between document-summary sentence pairs.
AlignScore~\cite{alignscore} extended this by training detection models on multiple semantic alignment tasks evaluated at the chunk level.
MiniCheck~\cite{minicheck} addressed data scarcity by synthesizing hallucinated examples using GPT-4 for model training.

The strong zero-shot instruction-following capabilities of modern LLMs have also enabled LLM-as-a-judge methods~\cite{llmjudge, luo2023chatgpt, jacovi2025facts, gao2023humanlikesummarizationevaluationchatgpt}.
Instead of evaluating entire generated summaries, approaches like FACTSCORE~\cite{min-etal-2023-factscore} and RAGAS~\cite{es-etal-2024-ragas} decompose summaries into claims for granular hallucination detection.

Like our \hallucinationleaderboard, efforts such as FACTS Grounding~\cite{jacovi2025facts} and Galileo’s Hallucination Index~\cite{GalileoAI_Hallucination_Index} also provide leaderboards to benchmark hallucinations in LLMs, relying on LLM judges employed in a zero-shot manner to evaluate responses.

LLM judges have also been employed to evaluate various aspects of RAG, including the relevance of retrieved passages~\cite{10.1145/3626772.3657707}, citation faithfulness~\cite{liu-etal-2023-evaluating}, and the factuality and completeness of responses~\cite{10.1145/3726302.3730090}.

Nonetheless, hallucination detection in RAG remains challenging, with weak effectiveness observed across current methods.
Benchmarks such as AggreFact, RAGTruth, TofuEval, and \faithbench consistently show limitations in existing methods, including LLM-based ones.
Notably, \faithbench highlighted that current methods, including using LLMs for classification, achieved near 50\% accuracy, suggesting negligible ability to identify hallucinated responses.
Further, both RAGTruth and TofuEval suggest that smaller, fine-tuned detection models can perform competitively with or even outperform zero-shot LLM-based evaluation approaches.

\section{Vectara's Original Hallucination Leaderboard}

In 2023, \Vectara's hallucination leaderboard was released using \Vectara's hallucination detection model, \hhem-1.0-open.
This model was later updated to \hhem-2.0 with improved effectiveness, the ability to handle longer contexts, and multilingual capabilities.
The current leaderboard relies on the open version, \hhem-2.1-open, publicly released on HuggingFace.
To date, \hhem has been downloaded over 4 million times, reflecting strong community interest and adoption.
While specific training details remain confidential, we note that \hhem-2.1-open was trained using the RAGTruth training set among other datasets.

\input{detection_evaluation}

To build our \hallucinationleaderboard, articles were selected from diverse sources such as BBC, CNN, Wikipedia, and the Daily Mail, following prior work on summarization evaluation and factuality verification~\cite{xsum, xsumfactuality, schuster-etal-2021-get, fever, summeval, polytope, frank, dailymailcnn}.
Articles containing objectionable content, which LLMs may refuse to summarize, were excluded.
The resulting dataset is composed of 955 articles with a median length of approximately 217 words.

LLMs are evaluated by prompting them to generate concise summaries strictly grounded on the provided passages.
\hhem then assesses the proportion of summaries generated by the LLM containing hallucinations.
Refusals are tracked by measuring the proportion of short responses (5 words or fewer).
Users are also invited to submit specific models for evaluation.
Continuously updated, the leaderboard now benchmarks hallucination rates of over 160 different LLMs, typically evaluating new models as soon as they become publicly available to track ongoing advances in LLMs.

\input{faithbench_detection}

\section{\faithbench}
\label{sec:faithbench_main}

\faithbench~\cite{bao-etal-2025-faithbench} examined hallucinations in diverse LLM-generated summaries and assessed the effectiveness of hallucination detection methods through human annotations.
It included summaries from ten state-of-the-art LLMs, including GPT-4o, GPT-3.5, Claude-3.5-Sonnet, Gemini-1.5-Flash~\cite{team2024gemini}, and open-source models like Llama-3.1~\cite{grattafiori2024llama}, revealing that hallucinations remain frequent and detection methods generally fail to identify them reliably or accurately.

Human annotators labelled hallucinations as Unwanted when the summary contained contradictory or unsupported information, Benign when the information was supported by world knowledge, but absent from the article, or Questionable when the classification was unclear.

Articles in \faithbench were selected based on frequent disagreements on summaries among hallucination detection models.
True-NLI, TrueTeacher, \hhem-2.1-open, and GPT-4o/GPT-3.5 judges using the chain-of-thought (CoT) prompt from \citet{luo2023chatgpt} were used to identify articles where summary hallucination classifications were most disagreed upon.
The dataset includes 75 articles, each with ten annotated summaries from different LLMs, allowing for many diverse LLM summaries to be studied per article. We show that this diversity in summaries also proves to be useful for \faithjudge when judging new summaries.

\section{\faithjudge}

Human annotation is the gold standard for hallucination detection, but it is time-consuming and expensive.
\faithjudge offers a scalable alternative by leveraging hallucination annotations to guide an LLM judge in evaluating new summaries.
We also expand \faithjudge to 
other RAG tasks, including question-answering (QA) and writing overviews from structured data in the JSON format using the RAGTruth dataset~\cite{ragtruth}.
This is detailed further in Section~\ref{ragtruth_faithjudge}.

To assess an LLM's response, \faithjudge involves prompting an LLM judge with other responses to the same prompt, along with their corresponding hallucination annotations.
These annotations include hallucination spans, source references from the context, and labels of either Benign, Unwanted, or Questionable, identified by multiple human annotators.

To evaluate the effectiveness of \faithjudge, we use the fact that each \faithbench article has summaries from ten different LLMs.
The judge is given the other nine annotated summaries as context, and its assessments on each summary from \faithbench are compared to human annotations.
As shown in Section~\ref{sec:results}, \faithjudge substantially improves automated hallucination evaluation, outperforming existing detection methods by leveraging human-labelled examples.
This allows for more accurate automated hallucination evaluation, where existing hallucination detection methods continue to lag.

\section{Evaluating Hallucination Detectors}
\label{sec:results}

\subsection{Evaluation Datasets}

We evaluate hallucination detection methods on four summarization datasets: \faithbench, AggreFact~\cite{aggrefact}, RAGTruth~\cite{ragtruth}, and TofuEval-MeetingBank~\cite{tang-etal-2024-tofueval}.
While each of these datasets has previously analyzed hallucination detection individually, we provide a comparison across all four, motivating the need for our \faithjudge approach.

For \faithbench, we assign each summary the most severe hallucination label given by a majority of the annotators.
We evaluate using summaries labelled either Unwanted or Consistent, excluding Benign and Questionable cases due to their more ambiguous nature.
This slightly differs from the original \faithbench evaluation, which pooled the worst label across all annotators for each summary and combined Benign cases with Consistent ones, while combining Unwanted cases with Questionable ones for the binary classification problem.

For AggreFact, we evaluate on the SOTA subset, which involves annotated summaries generated by fine-tuned T5~\cite{t5}, BART~\cite{lewis-etal-2020-bart}, and PEGASUS~\cite{pegasus} models.
For RAGTruth, we evaluate on the annotated summaries.
Lastly, for TofuEval, we evaluate on summaries generated using articles from the MeetingBank dataset~\cite{meetingbank}.

\input{labels_figure}

\subsection{Existing Hallucination Detectors}

Table~\ref{tab:organized-detection-models} compares the effectiveness of fine-tuned hallucination detectors and zero-shot LLM-based methods across various datasets.
We evaluate \Vectara's \hhem models alongside AlignScore, MiniCheck, including Bespoke-MiniCheck~\cite{tang2024bespokeminicheck}, and TrueTeacher.
We also include current LLMs, used in a zero-shot setting, such as GPT-4o and o3-mini-high, as well as open-source models Qwen2.5 (7B and 72B), Llama-3.1 (8B), and Llama-3.3 (70B).
The o3-mini model, in particular, excels in reasoning tasks.

Classification methods are separated into claim-wise and summary-wise classification.
Claim-wise evaluation involves decomposing sentences from summaries into individual claims using Llama-3.3 (70B) and a similar prompt from \citet{minicheck}, while summary-wise methods assess the entire summary at once.

For LLM-based detection, we test three prompts: (1) the RAGAS prompt~\cite{es-etal-2024-ragas}, which verifies lists of claims instead of entire responses, (2) the FACTS Grounding JSON prompt, shown to be the most effective of the prompts tested in ~\citet{jacovi2025facts} for GPT-4o, and (3) the CoT-based prompt from \citet{luo2023chatgpt}.
We modify prompts slightly as needed for clearer final outputs and to specifically evaluate summaries.

Table~\ref{tab:organized-detection-models} shows that, similar to findings in previous work, hallucination detection remains challenging.
Zero-shot classification using GPT-4o and o3-mini-high tends to perform best, both using summary-wise classification with either the FACTS Grounding JSON prompt or the \citet{luo2023chatgpt} prompt.
However, their average effectiveness remains modest, with balanced accuracy below 78\% and F1-macro below 72\%.
Considering \faithbench, the highest balanced accuracy is achieved by o3-mini-high at 68.8\% while the highest F1-macro of 63.7\% is achieved by the \hhem model when considering claim-wise classification.

\input{full_classification}

The table illustrates improved effectiveness with increased model size: larger open-source models generally outperform smaller ones, and GPT-4o and o3-mini-high achieve the highest overall effectiveness.
However, although \hhem-2.1-open is the smallest model tested, it performs strongly, outperforming several larger models.
Among the fine-tuned models, only the 7B-parameter MiniCheck achieves higher average scores for summary-wise classification, while both MiniCheck variants outperform it in claim-wise classification.

Overall, fine-tuned models can achieve stronger scores than smaller prompted LLMs, but the largest LLMs typically yield the best results, even while being zero-shot methods.
This suggests room for further improvement with LLM-based classification by learning from hallucination labels.
Regardless, the examined methods demonstrate modest effectiveness in general, with particularly weak effectiveness on \faithbench, which captures a diverse set of LLM summaries but is designed to be challenging for hallucination detection models.

\input{num_examples_figure}

\subsection{Evaluating \faithjudge}

Table~\ref{tab:faithbench-only} presents the effectiveness of \faithjudge on \faithbench using various LLMs.
The highest effectiveness is achieved using the o3-mini-high judge, reaching a balanced accuracy of 84\% and an F1-macro of 82.1\%, allowing for much higher agreement with human annotation on \faithbench than the existing hallucination detection methods discussed.
Although the effectiveness of \faithjudge is not perfect, this may be partly explained by disagreements in human annotation.
While human annotation is the gold standard, the \faithbench authors ~\cite{bao-etal-2025-faithbench} noted imperfect inter-annotator agreement in general and low inter-annotator agreement on more ambiguous and challenging to identify Benign and Questionable hallucination labels.

Effectiveness generally improves with increasing model size.
We also tested an ensemble approach, but found that combining predictions from multiple models, including Qwen2.5 (72B), Llama-3.3 (70B), and GPT-4o with a majority vote did not outperform o3-mini-high alone.
Therefore, we adopt the o3-mini-high judge as the standard for \faithjudge, with the possibility of using a stronger LLM judge in the future.

Figure~\ref{fig:predictions_per_model} displays the distribution of \faithjudge predictions across LLMs.
While effective, \faithjudge with o3-mini-high tends to underpredict hallucinations.
This is evident for Command-R, Mistral, and Qwen, where fewer summaries were flagged as hallucinated compared to the number labelled Unwanted by annotators in \faithbench.

\input{validating_faithjudge_ragtruth}

Table~\ref{tab:conf_matrices} presents confusion matrices for binary and ternary (including Benign labels) hallucination classification using \faithjudge.
We observe that Benign summaries are difficult for \faithjudge to classify correctly.
In the ternary setting, \faithjudge often misclassifies Benign summaries, generally labelling them as Consistent.
Similarly, Questionable summaries are classified unreliably, though this aligns with expectations.
For simplicity, we only employ \faithjudge for binary classification.

Figure~\ref{fig:num_examples} shows the sensitivity and specificity of \faithjudge as the number of annotated examples provided increases.
Specificity remains consistently high, though slightly decreasing as more examples are given, while sensitivity notably improves as the number of examples increases.
This indicates that providing more annotated examples leads \faithjudge to predict hallucinated cases more often and better identify hallucinations.

\section{Adding More Evaluation Tasks}
\label{ragtruth_faithjudge}

While \faithbench provides hallucination annotations across 10 different LLMs, it is limited to evaluating summaries only.
To broaden the scope of \faithjudge beyond summarization, we incorporate annotated responses from the RAGTruth dataset~\cite{ragtruth}.
RAGTruth includes three types of tasks: summarization, question-answering, and a data-to-text generation task that requires generating an overview of a business from JSON data sourced from the Yelp Open Dataset~\cite{yelp_open_dataset_2021}.
RAGTruth contains human-annotated hallucination labels for responses generated by six different LLMs: GPT-3.5, GPT-4~\cite{gpt4}, Llama-2 (7B, 13B, and 70B)~\cite{touvron2023llama}, and Mistral-7B~\cite{jiang2023mistral7b}.

For each RAGTruth task, we take up to 150 sources (articles for summarization, queries and passages for question-answering, and JSON data for data-to-text) with their corresponding annotated responses primarily from the test set and supplemented by the dev set where necessary.
We remove sources where none of the LLM responses have a hallucination annotation. 

Table~\ref{tab:comparison_facts_vs_faithjudge} compares the effectiveness of \faithjudge against the zero-shot FACTS Grounding JSON prompt, which was previously shown to be an effective prompt in \citet{jacovi2025facts}, on the \faithbench and RAGTruth subsets used in our leaderboard.
In each setting, \faithjudge achieves stronger agreement with human hallucination annotations, highlighting its strength across tasks beyond summarization.

\section{Conclusion}

In this paper, we presented our efforts at \Vectara in evaluating and benchmarking hallucinations in RAG, discussing and building on our established hallucination leaderboard, and proposing \faithjudge.
We identified effectiveness limitations in existing hallucination detection methods, including our own \hhem model.
To address these challenges, we proposed \faithjudge, an approach that leverages human hallucination annotations to enhance automated hallucination detection, achieving greater effectiveness, but requiring annotations from summaries of the same articles.

Our leaderboard is live on the \faithjudge GitHub, and we share some of these results in Appendix~\ref{appendix_rankings}, providing a framework for more accurate faithfulness evaluation across diverse models and RAG tasks.
We plan to continue to update our leaderboard to evaluate new models and to use improved LLM judges.

\section*{Limitations}

There are some limitations to our evaluation methodology.
First, our evaluation focuses exclusively on faithfulness and does not address the overall quality or helpfulness of summaries and answers.
Though summary and answer quality are important in RAG applications, we consider this evaluation largely orthogonal to faithfulness.

One issue to consider is that an extractive summarizer or an LLM that simply copies parts of the article or the entire article in its response would avoid hallucinations.
Nonetheless, we maintain that evaluating LLMs through hallucinations in generated summaries is promising because these hallucinations remain persistent in current LLMs.

Finally, while the o3-mini-high judge demonstrates strong effectiveness, there remains room for enhancing accuracy and agreement with human annotators.
We hope that as LLMs continue to improve, replacing o3-mini-high in \faithjudge may allow for more accurate and reliable evaluation.

\section*{Acknowledgements}

We respectfully acknowledge the late Simon Mark Hughes, who led the development of the original \hhem\ model and \Vectara's original hallucination leaderboard.
His contributions laid important groundwork for Vectara’s ongoing research and continue to leave a lasting influence on our work.

\bibliography{custom}

\clearpage

\appendix

\section*{Appendix}

\section{Summaries with FaithJudge Verdicts}

\input{judgement_examples}

Figure~\ref{fig:faithjudge_judgement_examples} shows examples of two different LLM summaries for a news source from FaithBench.
We generally observe that hallucinations can often be subtle and require great effort to identify.
For example, while the source identifies Paraxylene as a ``reportedly carcinogenic chemical'', Grok-3 drops the ``reportedly'' modifier and describes Paraxylene as a ``carcinogenic chemical''.
This removed detail is pertinent because it changes the information from the source.
Notably, to the best of our understanding, there is not enough evidence to say that Paraxylene is carcinogenic in humans or in animals~\cite{kandyala2010xylene}.

\input{rag_tasks}

\input{judge_bias}

\input{leaderboard-rankings}

\section{Judge Bias}

The FACTS Grounding leaderboard~\cite{jacovi2025facts} uses three different LLM judges to mitigate bias arising from any single judge favoring its own outputs.
Inspired by this, we analyze judge bias using Tables~\ref{tab:multi_judge_evaluation_results} and~\ref{tab:bias_totals_rankings}, which evaluate the impact of using different judges across all subsets included in our leaderboard.

Table~\ref{tab:multi_judge_evaluation_results} reports the effectiveness of three different LLMs when used as judges.
The table shows that o3-mini-high remains a relatively effective LLM for \faithjudge, often scoring the highest. 
The table also shows that while using multiple judges can improve effectiveness further, in some cases, individual LLMs can score higher than the majority vote approach between the three LLMs.
For example, o3-mini-high scores higher than the ensembling approach when evaluating on the RAGTruth QA subset.

Table~\ref{tab:bias_totals_rankings} explores how each judge model ranks the other LLMs.
Interestingly, {o3-mini-high} and {llama-4-maverick} both rank {gemini-2.0-flash} as having the fewest hallucinated responses, while {gemini-2.0-flash} ranks itself second to {o3-mini-high}, with only a small difference in hallucinated response counts (29 vs. 31).

While using multiple judges might enhance robustness and reduce individual model bias, we currently rely on a single judge to reduce computational costs.
As newer and stronger LLMs become available, we plan to update \faithjudge by substituting the current o3-mini-high judge model with a more effective one.

\section{\faithjudge Rankings}
\label{appendix_rankings}

Table~\ref{tab:hallucination_rates} presents \faithjudge rankings for a small set of LLMs.
In addition to detecting hallucinations, we also prompt \faithjudge to flag responses that are invalid, for example, when a model fails to meaningfully summarize an article.
For simplicity, since these cases are rare, we count these as hallucinated responses.
Models are ranked based on their overall hallucination rate, calculated as the total number of hallucinated or invalid responses across all four evaluation subsets.
We plan to continue evaluating LLMs using \faithjudge. Please see the current leaderboard, with many more LLMs evaluated, at \url{https://github.com/vectara/FaithJudge}.

\end{document}

%% file: detection_evaluation.tex
\begin{table*}[t]
  \centering

\resizebox{1.0\textwidth}{!}{
  \begin{tabular}{l|c|cc|cc|cc|cc||cc}
    \toprule
         & 
         & \multicolumn{2}{c}{{AggreFact-SOTA}} 
         & \multicolumn{2}{c}{{RAGTruth-Summ}} 
         & \multicolumn{2}{c}{{TofuEval-MB}} 
         & \multicolumn{2}{c}{{\faithbench}}
         & \multicolumn{2}{c}{\textbf{Average}} \\
    \cmidrule(lr){3-4}
    \cmidrule(lr){5-6}
    \cmidrule(lr){7-8}
    \cmidrule(lr){9-10}
    \cmidrule(lr){11-12}
    \textbf{Method} & \textbf{\# Params} 
      & Acc (\%) & F1 (\%) 
      & Acc (\%) & F1 (\%)
      & Acc (\%) & F1 (\%)
      & Acc (\%) & F1 (\%)
      & Acc (\%) & F1 (\%) \\
    \midrule
    \textbf{Claim-wise} & & & & & & & & & & & \\
      \quad \textit{Fine-Tuned Hallucination Detection Models} & & & & & & & & & & & \\
        \quad\quad \hhem-1.0-Open & 184M
             & 76.0 & 71.0
             & 66.2 & 52.2
             & 54.4 & 49.9
             & 59.3 & 58.8
             & 64.0 & 58.0 \\
        \quad\quad \hhem-2.1-Open & 110M
             & 73.2 & 69.7
             & 67.7 & 56.1
             & 60.9 & 61.2
             & 66.7* & 63.7*
              & 67.1 & 62.7 \\
        \quad\quad AlignScore-base & 125M
             & 69.5 & 61.9
             & 60.2 & 42.4
             & 51.7 & 44.0
             & 60.6 & 59.9
              & 60.5 & 52.1 \\
        \quad\quad AlignScore-large & 355M
             & 73.9 & 69.3
             & 67.9 & 54.1
             & 56.1 & 52.9
             & 62.8 & 59.6
             & 65.2 & 59.0 \\
        \quad\quad MiniCheck-Roberta-L & 355M
             & 75.7 & 72.5
             & 70.5 & 58.6
             & 67.6 & 68.5
             & 61.6 & 60.0
             & 68.8 & 64.9 \\
        \quad\quad Bespoke-MiniCheck & 7B
             & 74.3 & 70.1
             & 73.3 & 62.9
             & 76.9 & 78.4
             & 60.1 & 58.0
             & 71.2 & 67.3 \\
        \quad\quad TrueTeacher & 11B
             & 71.8 & 70.5
             & 56.5 & 56.1
             & 58.5 & 58.4
             & 59.8* & 51.8*
             & 61.7 & 59.2 \\
      \quad \textit{Zero-Shot Hallucination Detection with LLMs} & & & & & & & & & & & \\
      \quad\quad \textit{RAGAS Prompt} & & & & & & & & & & & \\
        \quad\quad\quad Qwen-2.5 & 7B
             & 71.1 & 69.0
             & 68.2 & 64.4
             & 64.3 & 57.7
             & 57.9 & 51.3
             & 65.4 & 60.6 \\
        \quad\quad\quad Qwen-2.5 & 72B
             & 74.4 & 69.9
             & 75.3 & 64.1
             & 69.2 & 70.6
             & 64.3 & 57.3
             & 70.8 & 65.5 \\
        \quad\quad\quad Llama-3.1 & 8B
             & 69.7 & 65.9
             & 68.5 & 59.9
             & 72.6 & 74.4
             & 60.3 & 57.9
             & 67.8 & 64.5 \\
        \quad\quad\quad Llama-3.3 & 70B
             & 77.3 & 74.9
             & 80.0 & 75.1
             & 73.2 & 70.6
             & 58.9 & 49.6
             & 72.3 & 67.5 \\
        \quad\quad\quad GPT-4o & ?
             & 75.9 & 70.3
             & 75.7 & 63.7
             & 75.2 & 76.7
             & 65.3 & 59.0
             & 73.0 & 67.4 \\
        \quad\quad\quad o3-mini-high & ?
             & 77.3 & 72.6
             & 74.6 & 62.9
             & 69.2 & 70.6
             & 67.4 & 60.7
             & 72.1 & 66.7 \\
    \midrule
    \textbf{Summary-wise} & & & & & & & & & & & \\
      \quad \textit{Fine-Tuned Hallucination Detection Models} & & & & & & & & & & & \\
        \quad\quad \hhem-1.0-Open & 184M
             & 78.9 & 79.7
             & 53.4 & 51.4
             & 56.5 & 39.8
             & 50.5 & 40.1
             & 59.8 & 52.7 \\
        \quad\quad \hhem-2.1-Open & 110M
             & 76.6 & 76.2
             & 64.4 & 67.1
             & 69.4 & 62.1
             & 52.6* & 32.9*
             & 65.8 & 59.6 \\
        \quad\quad AlignScore-base & 125M
             & 73.8 & 73.9
             & 57.6 & 58.2
             & 65.6 & 52.8
             & 51.3 & 33.8
             & 62.1 & 54.7 \\
        \quad\quad AlignScore-large & 355M
             & 72.7 & 74.2
             & 52.8 & 49.6
             & 57.4 & 39.2
             & 50.3 & 26.1
             & 58.3 & 47.3 \\
        \quad\quad MiniCheck-Roberta-L & 355M
             & 74.2 & 72.1
             & 66.3 & 60.9
             & 54.4 & 45.4
             & 55.0 & 53.2
             & 62.5 & 57.9 \\
        \quad\quad Bespoke-MiniCheck & 7B
             & 79.9 & 80.4
             & 79.4 & 77.1
             & 78.8 & 78.6
             & 55.7 & 47.3
             & 73.5 & 70.8 \\
        \quad\quad TrueTeacher & 11B
             & 77.6 & 78.4
             & 61.6 & 62.8
             & 57.4 & 39.2
             & 53.3* & 36.7*
             & 62.5 & 54.3 \\
        \quad \textit{Zero-Shot Hallucination Detection with LLMs} & & & & & & & & & & & \\
      \quad \quad \textit{FACTS Grounding Prompt} & & & & & & & & & & & \\
        \quad\quad\quad Qwen-2.5 & 7B
             & 66.9 & 68.7
             & 61.5 & 63.4
             & 62.8 & 54.4
             & 52.6 & 33.5
             & 60.9 & 55.0 \\
        \quad\quad\quad Qwen-2.5 & 72B
             & 71.6 & 73.7
             & 74.0 & 77.5
             & 68.8 & 58.8
             & 55.2 & 35.5
             & 67.4 & 61.4 \\
        \quad\quad\quad Llama-3.1 & 8B
             & 55.5 & 55.5
             & 62.9 & 62.7
             & 55.3 & 54.5
             & 60.9 & 49.7
             & 58.6 & 55.6 \\
        \quad\quad\quad Llama-3.3 & 70B
             & 79.3 & 78.1
             & 81.6 & 74.9
             & 70.1 & 71.3
             & 66.6 & 58.4
             & 74.4 & 70.7 \\
        \quad\quad\quad GPT-4o & ?
             & 81.6 & 78.8
             & 82.6 & 76.6
             & 76.3 & 76.0
             & 65.9 & 56.2
             & 76.6 & \textbf{71.9} \\
        \quad\quad\quad o3-mini-high & ?
             & 82.1 & 77.8
             & 79.8 & 70.6
             & 69.2 & 70.6
             & 68.8 & 60.7
             & 75.0 & 69.9 \\
             
        \quad\quad \textit{~\citeauthor{luo2023chatgpt} Prompt} & & & & & & & & & & & \\
        \quad\quad\quad Qwen-2.5 & 7B
             & 72.8 & 73.5
             & 67.6 & 70.2
             & 69.0 & 66.3
             & 53.4 & 39.0
             & 65.7 & 62.2 \\
        \quad\quad\quad Qwen-2.5 & 72B
             & 78.4 & 78.0
             & 81.3 & 81.1
             & 83.4 & 80.0
             & 58.3 & 44.3
             & 75.3 & 70.8 \\
        \quad\quad\quad Llama-3.1 & 8B
             & 60.8 & 51.2
             & 63.7 & 52.1
             & 57.1 & 55.8
             & 51.3 & 51.0
             & 58.2 & 52.5 \\
        \quad\quad\quad Llama-3.3 & 70B
             & 79.2 & 79.1
             & 81.3 & 82.9
             & 73.6 & 66.5
             & 58.8 & 43.6
             & 73.2 & 68.0 \\
        \quad\quad\quad GPT-4o & ?
             & 80.4 & 77.5
             & 85.1 & 80.9
             & 81.6 & 78.7
             & 62.5* & 50.6*
             & \textbf{77.4} & \textbf{71.9} \\
        \quad\quad\quad o3-mini-high & ?
             & 82.6 & 80.9
             & 83.2 & 80.6
             & 75.6 & 73.7
             & 63.3 & 49.8
             & 76.2 & 71.2 \\
    \bottomrule
  \end{tabular}
  }

  \caption{\small Balanced Accuracy and F1-Macro of hallucination detection methods across four datasets. 
  The final two columns report the simple average across the four datasets.
  We note that certain models marked with an asterisk (*) were used to select articles for the adversarially challenging \faithbench dataset.
Consequently, these models, including \hhem, may perform worse on \faithbench in summary-wise classification than they otherwise would.}


  \label{tab:organized-detection-models}
\end{table*}

%% file: faithbench_detection.tex
\begin{table*}[t]
  \centering
    \resizebox{0.72\textwidth}{!}{
  \begin{tabular}{l c|cc}
    \toprule
         & 
         & \multicolumn{2}{c}{{\faithbench}} \\
    \cmidrule(lr){3-4}
    \textbf{Method} & \textbf{\# Params} 
      & Acc (\%) & F1 (\%) \\
    \midrule

    \textit{FACTS Grounding Prompt} & & & \\
        \quad GPT-4o & ?
             & 65.9 & 56.2 \\
        \quad o3-mini-high & ?
             & 68.8 & 60.7 \\ 
    \textit{~\citeauthor{luo2023chatgpt} Prompt} & & & \\
        \quad GPT-4o & ?
             & 62.5 & 50.6 \\
        \quad o3-mini-high & ?
             & 63.3 & 49.8 \\
             
    \textit{\faithjudge Prompting} & & & \\
         \quad Qwen-2.5 & 7B
               & 71.9 & 66.6 \\
         \quad Qwen-2.5 & 72B
               & 73.2 & 73.0 \\
         \quad Llama-3.1 & 8B
               & 60.8 & 61.0 \\
         \quad Llama-3.3 & 70B
               & 77.5 & 77.8 \\
         \quad GPT-4o & ?
               & 79.5 & 81.1 \\
         \quad o3-mini-high & ?
               & \textbf{84.0} & \textbf{82.1} \\
        \quad Majority Vote (Qwen 72B, Llama 70B, GPT-4o) &  
               & 80.7 & 81.3 \\

    \bottomrule
  \end{tabular}
  }
  \caption{\small Balanced Accuracy and F1-Macro for \faithjudge on \faithbench using different LLM judges. When combining models, we apply majority voting and break ties by defaulting to a classification of Inconsistent.}
  \label{tab:faithbench-only}
\end{table*}

%% file: labels_figure.tex
\begin{figure*}[t]
    \centering
 \includegraphics[width=0.82\textwidth]{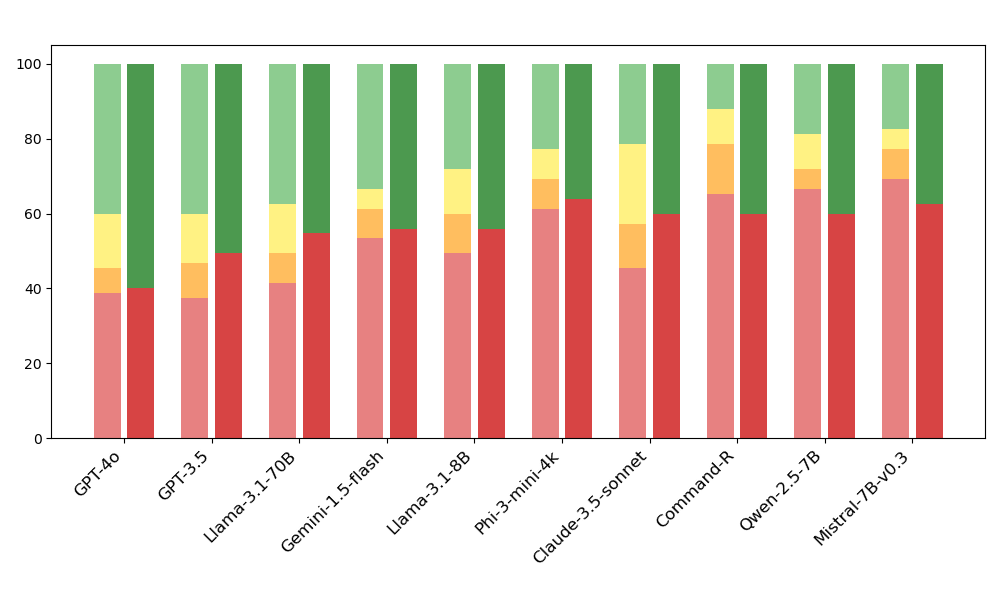}
    \caption{Proportion of summary \faithbench labels (left) and \faithjudge predictions (right) across models. For \faithbench labels, red indicates Unwanted, orange indicates Questionable, yellow indicates Benign, while green indicates Consistent. For \faithjudge predictions, red indicates Hallucinated, and green indicates Consistent summaries. Each bar shows the proportion of summaries falling into each category.}
    \label{fig:predictions_per_model}

\end{figure*}

%% file: full_classification.tex
\begin{table}[t]
\centering

\resizebox{1.0\columnwidth}{!}{%
  \begin{tabular}{c}
    \begin{tabular}{r|rr}
      \multicolumn{3}{c}{\textbf{Binary Classification}} \\ \hline
      \textbf{Gold Truth} & \textbf{Consistent} & \textbf{Inconsistent} \\ \hline
      Unwanted     & $74$ & $322$ \\
      Questionable & $29$ & $38$ \\
      Benign       & $50$ & $34$ \\
      Consistent   & $176$ & $27$ \\
    \end{tabular}
    \\[1ex]
    \begin{tabular}{r|rrr}
      \multicolumn{4}{c}{\textbf{Ternary Classification}} \\ \hline
      \textbf{Gold Truth} & \textbf{Consistent} & \textbf{Benign} & \textbf{Unwanted} \\ \hline
      Unwanted     & $84$ & $18$ & $294$ \\
      Questionable & $28$ & $13$ & $26$ \\
      Benign       & $51$ & $10$ & $23$ \\
      Consistent   & $179$ & $4$ & $20$ \\
    \end{tabular}
  \end{tabular}
}
\caption{Confusion matrices for \faithjudge prompted for classification on \faithbench summaries.}
\label{tab:conf_matrices}

\end{table}

%% file: num_examples_figure.tex
\begin{figure}[t]
    \centering    \includegraphics[width=1.0\columnwidth]{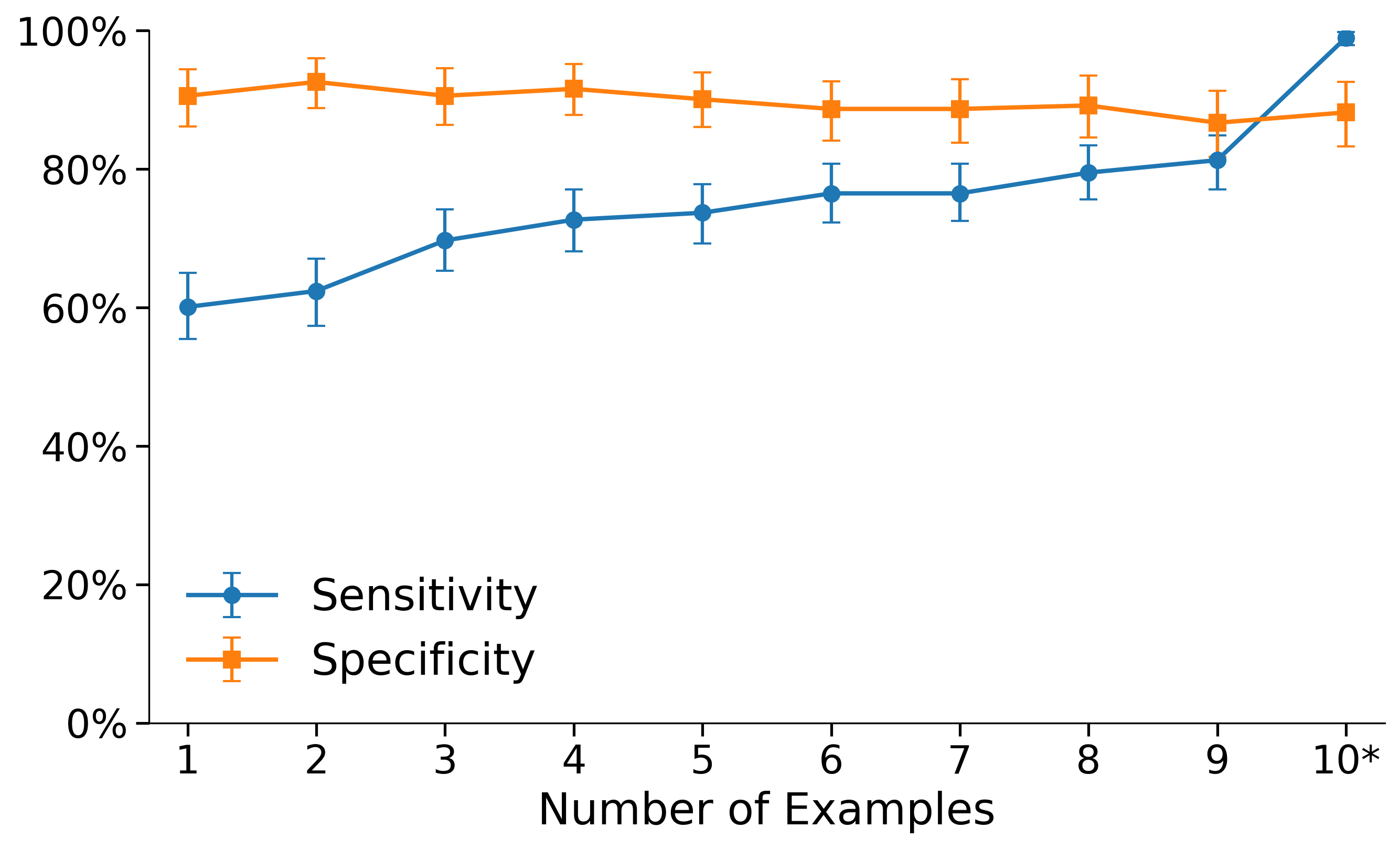}
    \caption{Sensitivity and specificity with \faithjudge as the number of examples in the prompt is increased.
    We place an asterisk (*) next to the 10 because, in this case, \faithjudge is shown annotations for the summary it is evaluating.}
\label{fig:num_examples}
\end{figure}

%% file: validating_faithjudge_ragtruth.tex
\begin{table*}[ht!]
\centering
\resizebox{0.8\textwidth}{!}{%
\begin{tabular}{l cc cc}
\toprule
\multirow{2}{*}{\textbf{Dataset}} & \multicolumn{2}{c}{\textbf{FACTS Grounding Prompt}} & \multicolumn{2}{c}{\textbf{FaithJudge Prompt}} \\
\cmidrule(lr){2-3} \cmidrule(lr){4-5}
                         & F1-Macro & Balanced Accuracy & F1-Macro & Balanced Accuracy \\
\midrule
RAGTruth-Data-to-text       & 77.1   & 75.1            & 86.3   & 85.1            \\
RAGTruth-QA             & 76.9   & 81.6            & 83.4   & 85.4            \\
RAGTruth-Summary        & 73.6   & 80.3            & 80.2   & 84.9            \\
FaithBench-Summary      & 54.3   & 65.2            & 70.8   & 77.6            \\
\bottomrule
\end{tabular}
}
\caption{Comparison between the FACTS Grounding zero-shot prompting approach (JSON Prompt) and the \faithjudge prompting approach on the subsets of data used in our leaderboard.
In all cases we use an o3-mini-high LLM judge. For \faithjudge, we prompt the judge to evaluate LLM responses by providing the responses from the other LLMs in the dataset with their corresponding annotations.
For \faithbench, we evaluate using all summaries, treating summaries labelled as Questionable or Benign as inconsistent summaries. }
\label{tab:comparison_facts_vs_faithjudge}
\end{table*}

%% file: judgement_examples.tex
\definecolor{softgray}{RGB}{246,246,246}
\tcbset{
  colback=white,
  colframe=black!10,
  boxrule=0.5pt,
  arc=2pt,
  left=8pt,right=8pt,top=8pt,bottom=8pt,
  enhanced jigsaw,
}

\newcommand{\VerdictConsistent}{\colorbox{green!15}{\textsf{\textcolor{green!40!black}{Consistent}}}}
\newcommand{\VerdictInconsistent}{\colorbox{red!15}{\textsf{\textcolor{red!60!black}{Inconsistent}}}}

\newcommand{\ModelCard}[4]{%
\begin{tcolorbox}[title={\textbf{#1} \hfill Verdict: #2}]
\textbf{Summary}\par
\small #3

\vspace{0.75em}
\textbf{FaithJudge:}\par
\small #4
\end{tcolorbox}
}

\newtcbox{\SourceTag}{on line, arc=2pt, boxrule=0.5pt,
  colback=black!15, colframe=black!35, coltext=black,
  left=7pt, right=7pt, top=2pt, bottom=2pt,
  fontupper=\sffamily\bfseries\footnotesize}

\begin{figure*}[ht!] 
\centering

\begin{tcolorbox}[breakable,
  title=,
  title filled=true, colbacktitle=black!12, coltitle=black,
  colback=white, colframe=black!10, boxrule=0.5pt, arc=2pt,
  left=8pt, right=8pt, top=8pt, bottom=8pt, boxsep=5pt,
  before skip=0pt, after skip=0pt
]

\small

\SourceTag{\textsc{Source}}\par
\vspace{0.2cm}
{\color{black!80}
hong kong ( cnn ) six people were hurt after an explosion at a controversial chemical plant in china 's southeastern fujian province sparked a huge fire , provincial authorities told state media.
the plant , located in zhangzhou city , produces paraxylene ( px ) , a reportedly carcinogenic chemical used in the production of polyester films and fabrics~ .
the blast occurred at an oil storage facility monday night after an oil leak , though local media has not reported any toxic chemical spill~.
five out of six people were injured by broken glass and have been sent to the hospital for treatment , xinhua news agency reported~.
more than 600 firefighters were sent to battle the blaze and the fire is now under control , the zhangzhou fire department wrote on their official microblogging account tuesday morning~.
residents living close to the plant had heard the explosion and took to weibo to post photos of the fire~.
one user wrote that he heard a loud blast and felt slight tremors~.
the plant was hit by another explosion in july 2013 , although there were no reports of casualties or toxic leaks at the time~.
though demonstrations are illegal in china , the construction of px plants has sparked protests , which have occasionally turned violent , in several cities in recent years~.
the zhangzhou plant was slated for xiamen -- a densely populated city in the southeast of the country~.
however , it provoked an angry backlash in 2007 due to pollution concerns and prompted the local government to relocate the factory to its current , more remote location~.
}

\medskip\hrule\medskip

\textbf{Gemini-2.5-Pro} \hfill \textbf{Verdict: } \VerdictInconsistent\par

\medskip
{\textcolor{black!60}{Summary}}\par
{\color{black!80}
An explosion at a chemical plant in Zhangzhou, China, injured six people and caused a large fire. The plant produces paraxylene (PX), a reportedly carcinogenic chemical. The blast occurred on a Monday night at an oil storage facility after an oil leak, though no toxic chemical spill has been reported. Five of the six people were injured by broken glass. Over 600 firefighters were sent to battle the blaze, which is now under control. \textcolor{red}{The plant had a previous explosion in July 2013 and was relocated to its current location from Xiamen} in 2007 due to \textcolor{red}{public protests} over pollution concerns.
}

\medskip
{\textcolor{black!60}{FaithJudge Response}\par
{\color{black!80}
The summary accurately reflects almost all details from the article. It correctly states that the explosion injured six people, that five of them were injured by broken glass, that the blast occurred at an oil storage facility on Monday night after an oil leak, that no toxic chemical spill was reported, that over 600 firefighters controlled the fire, and that there was a previous explosion in July 2013. 

However, the summary’s last sentence introduces a key hallucination. Instead of stating that the plant ``was slated for Xiamen'' and that it ``provoked an angry backlash in 2007 due to pollution concerns'' (which then led the government to relocate the factory), the summary claims it ``was relocated to its current location from Xiamen in 2007 due to public protests over pollution concerns.'' This rephrasing mischaracterizes the source in two ways: 

1. It implies the plant was actually located in Xiamen before being moved, rather than originally being planned there.

2. It changes ``an angry backlash'' to ``public protests,'' which is not how the source describes the reaction.

Because these details are not supported by the source text, the summary contains hallucinations.

Final classification: Inconsistent
}

\medskip\hrule\medskip

\textbf{Grok-3} \hfill \textbf{Verdict: } \VerdictInconsistent\par
\medskip
{\textcolor{black!60}{Summary}}\par
{\color{black!80}
An explosion at a chemical plant producing \textcolor{red}{paraxylene (PX), a carcinogenic chemical} used in polyester production, in Zhangzhou city, Fujian province, China, injured six people, five of whom were hurt by broken glass and hospitalized. The blast, caused by an oil leak at an oil storage facility, occurred on Monday night and led to a massive fire, which was controlled by over 600 firefighters by Tuesday morning. No toxic chemical spill was reported. Local residents heard the explosion, felt tremors, and shared images of the fire on Weibo. This plant, previously \textcolor{red}{the site of a 2013 explosion with no casualties or leaks,} \textcolor{red}{was relocated from Xiamen} to its current remote location due to pollution concerns and public backlash in 2007. PX plants have been controversial in China, sparking protests, sometimes violent, despite demonstrations being illegal.
}

\medskip
{\textcolor{black!60}{FaithJudge Response}}\par
{\color{black!80}
This summary introduces several details that do not exactly match the source. For example, it describes PX as ``a carcinogenic chemical'' instead of ``a reportedly carcinogenic chemical,'' changing the certainty level. It also states that the plant ``was relocated from Xiamen'' rather than noting that it was originally slated for Xiamen before the backlash prompted relocation. In addition, the summary phrases the 2013 explosion as having ``no casualties or leaks'' rather than saying there were ``no reports'' of casualties or toxic leaks. These deviations, even if minor, represent hallucinated details.

Final classification: Inconsistent
}}

\end{tcolorbox}

\caption{Two LLM summaries of a news source from FaithBench with FaithJudge hallucination judgements. The red text indicates portions of the summaries that contain hallucinations.}
\label{fig:faithjudge_judgement_examples}
\end{figure*}

%% file: rag_tasks.tex
\begin{table*}[ht!]
\centering
\resizebox{0.68\textwidth}{!}{%

\begin{tabular}{llcc}
\toprule
Dataset & Judge Model & F1-Macro & Balanced Accuracy \\
\midrule
\multirow{4}{*}{RAGTruth (Data-to-text)} 
    & o3-mini-high         & \underline{86.3} & \underline{85.1} \\
    & gemini-2.0-flash & 83.6 & 84.0 \\
    & llama-4-maverick     & 82.1 & 80.6 \\
    & \textit{Majority Vote}             & \textbf{86.4} & \textbf{85.8} \\
\midrule
\multirow{4}{*}{RAGTruth (QA)}
    & o3-mini-high         & \underline{\textbf{83.4}} & \underline{\textbf{85.4}} \\
    & gemini-2.0-flash & 81.8 & 84.2  \\
    & llama-4-maverick     & 77.5 & 81.2  \\
    & \textit{Majority Vote}        & 81.0 & 83.8  \\
\midrule
\multirow{4}{*}{RAGTruth (Summary)}
    & o3-mini-high         & 80.2 & \underline{84.9} \\
    & gemini-2.0-flash & \underline{83.6} & 82.7 \\
    & llama-4-maverick     & 78.0 & 83.7 \\
    & \textit{Majority Vote}             & \textbf{84.6} & \textbf{88.0} \\
\midrule
\multirow{4}{*}{FaithBench (Summary)}
    & o3-mini-high         & 70.8 & \underline{77.6} \\
    & gemini-2.0-flash & 66.1 & 75.5 \\
    & llama-4-maverick     & \textbf{\underline{74.7}} & 76.9 \\
    & \textit{Majority Vote}             & 72.4 & \textbf{79.1} \\
\bottomrule
\end{tabular}
}
\caption{Evaluation results for three models and an ensemble approach on the subsets of data used in our leaderboard. 
Here, for \faithbench, we evaluate using all summaries, treating summaries labelled as Questionable or Benign as inconsistent summaries.}
\label{tab:multi_judge_evaluation_results}
\end{table*}

%% file: judge_bias.tex
\begin{table*}[ht!]
\centering

\resizebox{\textwidth}{!}{
\begin{tabular}{l|cc|cc|cc}
\toprule
 & \multicolumn{2}{c}{Judged by o3-mini-high} & \multicolumn{2}{c}{Judged by gemini-2.0-flash} & \multicolumn{2}{c}{Judged by llama-4-maverick} \\
\cmidrule(r){2-3} \cmidrule(r){4-5} \cmidrule(r){6-7}
Evaluated Model & Hallucinated Responses & Rank & Hallucinated Responses & Rank & Hallucinated Responses & Rank \\
\midrule
gemini-2.0-flash    &  52 & 1 & 31  & 2 & 71  & 1 \\
o3-mini-high        &  64 & 2 & 29  & 1 & 94  & 2 \\
llama-4-maverick    & 105 & 3 & 72  & 3 & 110 & 3 \\
\bottomrule
\end{tabular}%
}
\caption{Total number of hallucinated responses per evaluated model, as judged by each model. Rankings indicate relative effectiveness in terms of hallucination frequency, from least to most.}
\label{tab:bias_totals_rankings}

\end{table*}

%% file: leaderboard-rankings.tex
\begin{table*}[ht!]
  \centering
  \resizebox{\textwidth}{!}{%
    \begin{tabular}{rlccccc}
      \toprule
      Rank & Model               & Overall Hallucination Rate & FaithBench (Summary) & RAGTruth (Summary) & RAGTruth (QA) & RAGTruth (Data-to-Text) \\
      \midrule
      1    & gemini-2.5-pro      & 6.65\%        & 18/72      & 7/150         & 2/139         & 7/150                   \\
      2    & gemini-2.0-flash    & 10.18\%       & 21/72      & 10/150         & 1/139         & 20/150                  \\
      3    & gpt-4.5-preview     & 11.94\%       & 27/72      & 15/150         & 7/139         & 12/150                  \\
      4    & o3-mini-high        & 12.52\%       & 25/72      & 12/150         & 9/139         & 18/150                  \\
      5    & gpt-3.5-turbo       & 14.87\%       & 32/72      & 13/150         & 8/139         & 23/150                  \\
      6    & gpt-4o              & 15.85\%       & 29/72      & 15/150         & 7/139         & 30/150                  \\
      7    & claude-3.7-sonnet   & 16.05\%       & 28/72      & 22/150         & 13/139        & 19/150                  \\
      8    & llama-3.3-70b       & 16.44\%       & 32/72      & 13/150         & 6/139         & 33/150                  \\
      9    & phi-4               & 17.03\%       & 32/72      & 12/150         & 6/139         & 37/150                  \\
      10   & mistral-small-24b   & 17.03\%       & 31/72      & 15/150         & 14/139        & 27/150                  \\
      11   & llama-4-maverick   & 20.55\%       & 37/72      & 20/150         & 13/139        & 35/150                  \\
      12   & llama-3.1-8b       & 28.38\%       & 32/72      & 19/150         &
      17/139       & 77/150                   \\
      \bottomrule
    \end{tabular}%
  }
  \caption{\faithjudge rankings for 12 LLMs, based on the number of hallucinated responses across four evaluation subsets: article summarization from \faithbench and RAGTruth, as well as question answering and data-to-text writing from RAGTruth.}
  \label{tab:hallucination_rates}
\end{table*}